\documentclass{article}
\usepackage{spconf,amsmath,graphicx}
\usepackage{multirow}
\usepackage{url}
\usepackage{amsmath, amssymb}
\usepackage{type1cm}
\usepackage{xcolor}

\usepackage[breaklinks=true,bookmarks=false,colorlinks,linkcolor=blue,citecolor=orange]{hyperref}

\title{Multi-Object Tracking as Attention Mechanism}
\name{Hiroshi Fukui\dag, Taiki Miyagawa\dag, Yusuke Morishita\dag}
\address{\dag NEC Corporation, Japan}
\begin{document}
%
\maketitle
\begin{abstract}
We propose a conceptually simple and thus fast multi-object tracking (MOT) model that does not require any attached modules, such as the Kalman filter, Hungarian algorithm, transformer blocks, or graph networks.
Conventional MOT models are built upon the multi-step modules listed above, and thus the computational cost is high.
Our proposed end-to-end MOT model, \textit{TicrossNet}, is composed of a base detector and a cross-attention module only. 
As a result, the overhead of tracking does not increase significantly even when the number of instances ($N_t$) increases.
We show that TicrossNet runs \textit{in real-time}; specifically, it achieves 32.6 FPS on MOT17 and 31.0 FPS on MOT20 (Tesla V100), which includes as many as $>$100 instances per frame. We also demonstrate that TicrossNet is robust to $N_t$; thus, it does not have to change the size of the base detector, depending on $N_t$, as is often done by other models for real-time processing.
\footnote{© 2023 IEEE. Personal use of this material is permitted. Permission from IEEE must be obtained for all other uses, in any current or future media, including reprinting/republishing this material for advertising or promotional purposes, creating new collective works, for resale or redistribution to servers or lists, or reuse of any copyrighted component of this work in other works.}
\end{abstract}
\section{Introduction}
\label{sec:intro}
Simple and fast models for multi-object tracking (MOT) have been studied to realize real-time tracking.
In particular, single-shot models have remarkably improved the tracking speed by linking detection and tracking in an end-to-end trainable manner~\cite{Yifu2020,Xingyi2020,Yifu2022,Zeng2022}.
Many such models are composed of three key processes: detection or feature extraction, re-identification (reID), and trajectory refinement \cite{Yifu2020, Zeng2022, Peize2020,Nicolai2017}.
To construct such a model, several attached modules are generally used, such as the Kalman filter, Hungarian algorithm, cosine similarity, boxIoU, transformer blocks, or graph networks \cite{Long2018,Xingyi2020,Yihong2021,Yifu2020,Peize2020,Zeng2022,Yifu2022}.
These attached modules increase computational cost especially when the number of tracking instances is large \cite{Yifu2020, Yifu2022}.
Therefore, such a model design is inefficient in this case.

\begin{figure}[t]
  \begin{center}
  \includegraphics[width=0.9\linewidth]{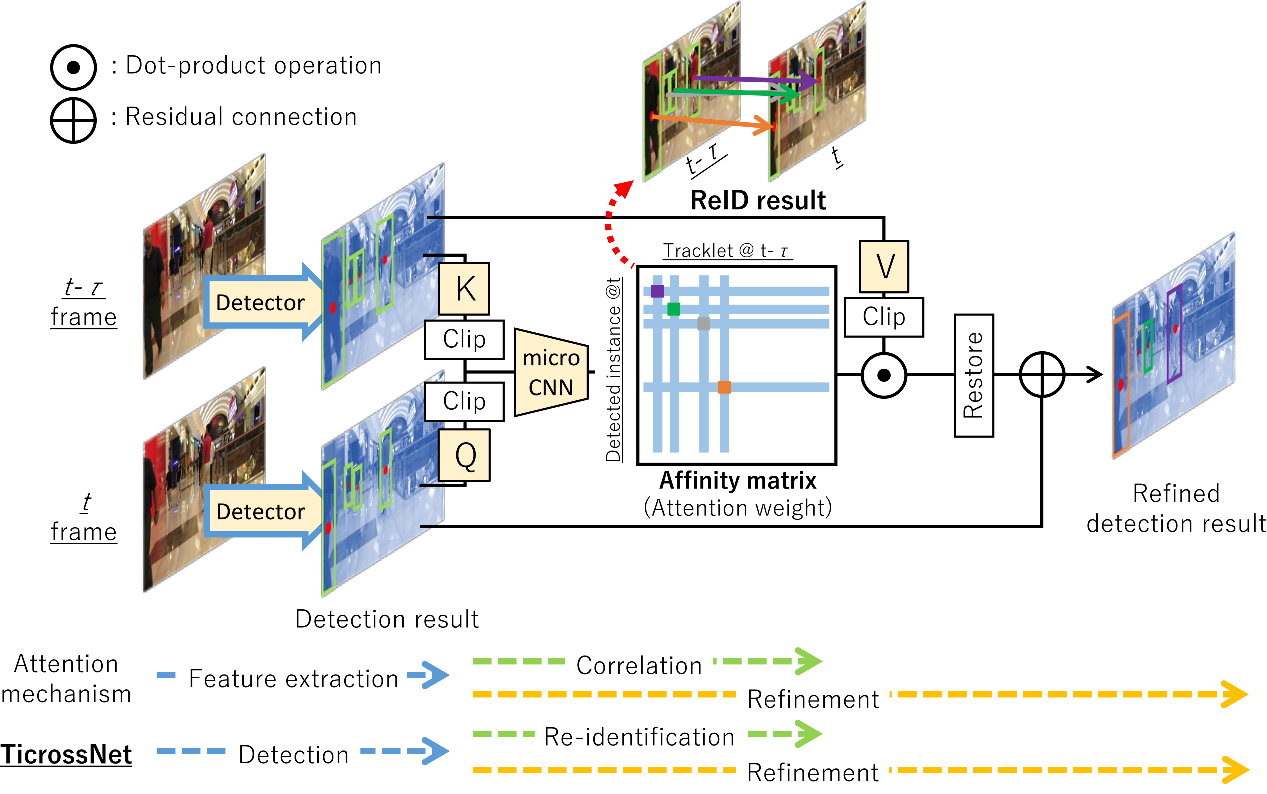}
  \caption{{\it TicrossNet} for online MOT. It consists of a base detector and a single cross-attention module only and thus is conceptually simple and fast. See Section \ref{sec:proposed} and Fig. \ref{fig:detailview} for details.}
  \label{fig:overview}
  \end{center}
\end{figure}

To make an efficient model for MOT, we attempt to reduce the number of modules and module complexity.
Our core idea is to use the cross-attention mechanism~\cite{Ashish2017} for MOT modeling. 
This idea is based on the similarity between the cross-attention mechanism and the key processes of MOT (Tab. \ref{tab:attn_vs_mot}).
We argue that this similarity enables us to perform MOT using only one cross-attention module (and a base detector).
As a result, we can complete all the key processes of MOT in GPU only unlike conventional models (except for MOTR \cite{Zeng2022}).

\begin{table*}
  \centering
  \scalebox{0.9}{
  \begin{tabular}{c|p{4.5cm}p{4.5cm}p{4.5cm}}
    \hline 
    Key process & Feature extraction & Correlation & Refinement \\
    \hline
    MOT & Feature extraction using {\it DNNs} & Estimating correlations between {\it instances} and {\it tracklets} & Refining {\it tracking trajectories} using reID results \\
    Attention & Feature extraction using {\it attention module} & Estimating correlations between {\it tokens} & Refining {\it input features} using feature correlation \\ \hline
  \end{tabular}
  }
  \caption{The similarity of each key process in MOT and attention mechanism. MOT computes the reID result using the output of the detector or backbone and refines a trajectory using the reID result. 
Similarly, the attention mechanism computes the attention weight using the feature vector in the previous layer and refines the input feature vector. }
  \label{tab:attn_vs_mot}
\end{table*}

Following the idea above, we propose a conceptually simple and fast tracker, called tracking crossword network (\textit{TicrossNet}), which completes all the MOT key processes using the cross-attention mechanism.
It requires only minor modifications to the vanilla cross-attention mechanism, i.e., a softmax normalization, feature clipping, and micro convolutional neural network (CNN)~(Fig.~\ref{fig:overview}), which do not increase computational cost significantly.
As a result, the overhead of the tracking process does not increase significantly even when the number of instances increases.
Note that TicrossNet uses the cross-attention mechanism for efficient MOT modeling, not only for feature extraction like conventional transformers~\cite{Zeng2022,Peize2020,Yihong2021}.
Note also that TicrossNet uses only one single-head cross-attention module unlike MOTR \cite{Zeng2022}.

Our experimental results show that TicrossNet achieves 32.6 FPS on MOT17 \cite{Anton2016} and 31.0 FPS on MOT20 \cite{Patrick2020} (using Tesla V100) even though the latter includes as many as $>$100 instances per frame. 
Note that the video frame rates of MOT17 and MOT20 are 30 and 25 FPS, respectively.
Therefore, we can safely say that TicrossNet runs \textit{in real-time}.
Furthermore, we show that TicrossNet maintains the processing speed even when the number of instances increases, while competitive baseline models~\cite{Yifu2020,Xingyi2020,Yifu2022} slow down significantly.

\section{TicrossNet}
\label{sec:proposed}

\subsection{Architecture} 
\label{sec:architecture}
\begin{figure}[t]
  \begin{center}
  \includegraphics[width=1.0\linewidth]{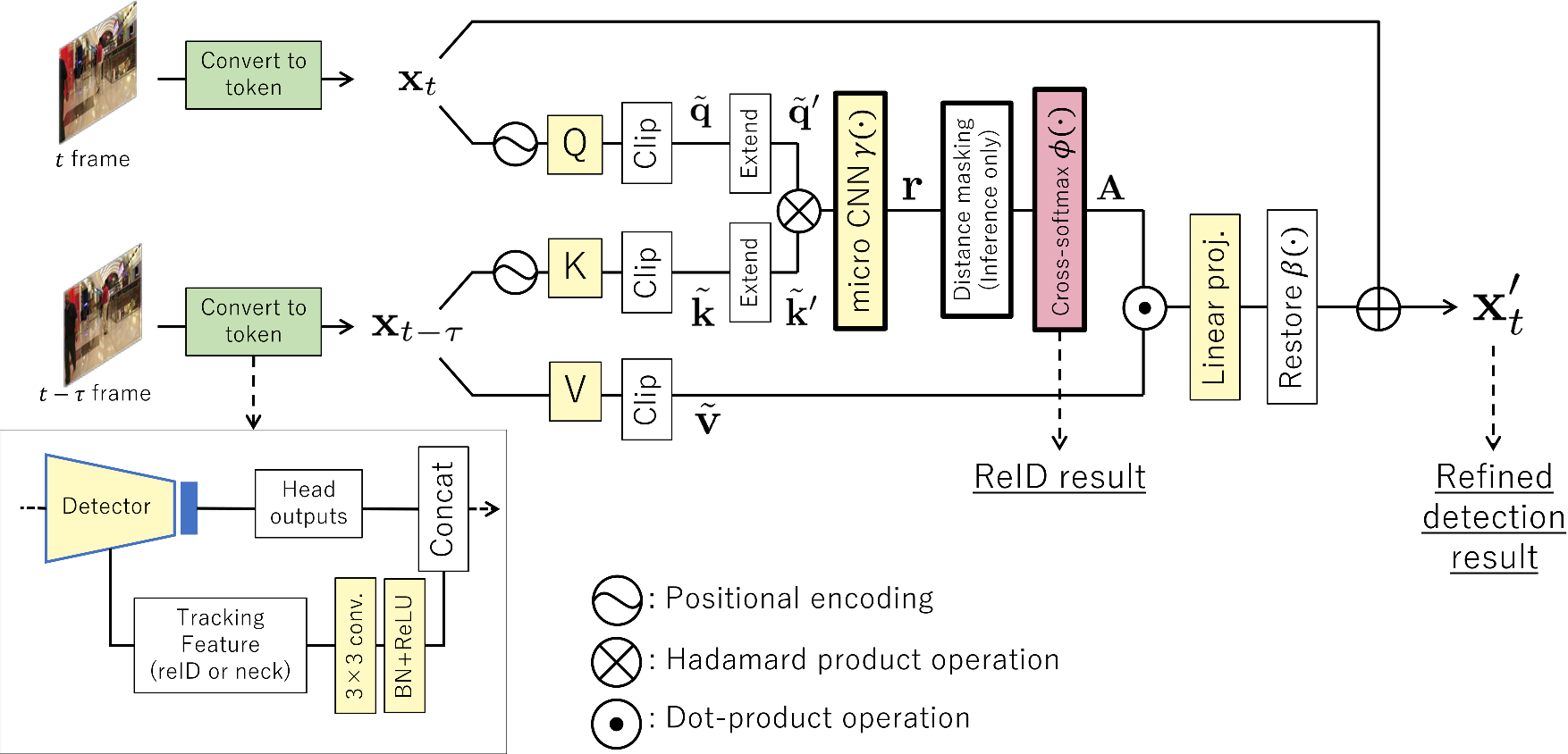}
  \caption{The detailed structure of TicrossNet. See Section \ref{sec:proposed}.}
  \label{fig:detailview}
  \end{center}
\end{figure}

The proposed tracking process is formally summarized as:
\begin{gather}
  \label{eq:pe}
  {\bf x}_{t, +}=pe({\bf x}_t), \,\, {\bf x}_{t-\tau, +}=pe({\bf x}_{t-\tau}),  \\
  \label{eq:def_embedding}
  {\bf q}=Q({\bf x}_{t,+}), {\bf k}=K({\bf x}_{t-\tau,+}), {\bf v}=V({\bf x}_{t-\tau}), \\
  \label{eq:feature_clipping}
  \tilde{\bf q}=clip({\bf q}), \,\, \tilde{\bf k}=clip({\bf k}), \,\, \tilde{\bf v}=clip({\bf v}),
  \\
  \label{eq:extend}
  \tilde{\bf q}'=extend(\tilde{{\bf q}}), \,\, \tilde{\bf k}'=extend(\tilde{{\bf k}}),\\
  \label{eq:attention}
  {\bf A} = attention(\tilde{{\bf q}}', \,\,\tilde{{\bf k}}', \,\, \tilde{{\bf v}}) = \phi \left(\gamma  \left(\tilde{{\bf q}}' \otimes \tilde{{\bf k}}'^{\top} \right)\right) \odot \tilde{{\bf v}} ,  \\
  \label{eq:tix_attention}
   {\bf x}'_{t} = \beta \left( {\bf A} \odot {\bf W}_{p} \right) +  {\bf x}_{t} ,
\end{gather}
where $\odot$ and $\otimes$ are the dot product and the Hadamard product operations, respectively. ${\bf W}_{p} \in \mathbb{R}^{D \times D}$ is a trainable matrix for the linear projection layer, where
$D$ is the sum of the numbers of channels in the output layer of the detector.
In the following, we explain the details of Eqs. \ref{eq:pe}--\ref{eq:tix_attention} and Fig. \ref{fig:detailview} in a step-by-step manner. The point is Eqs. \ref{eq:attention} \& \ref{eq:tix_attention} (${\bf A}$ and ${\bf x}'_{t}$).

\hspace{-6mm} {\bf 1: Detector.}
TicrossNet starts from a base detector of the instances in each frame (Figs.~\ref{fig:overview} \& \ref{fig:detailview}). 
We use CenterNet~\cite{Xingyi2019}, which is a commonly used detector for MOT because of its good tradeoff between speed and accuracy~\cite{Yifu2020,Xingyi2020,Yifu2022}.
We also use the reID feature (Fig. \ref{fig:detailview}), as is also done in FairMOT~\cite{Yifu2020}, to boost the accuracy of MOT.

\hspace{-6mm} {\bf 2: Input token~${\bf x}_t$.}
The detector's output is then used to make the input token at times $t-\tau$ and $t$, which are denoted by ${\bf x}_{t-\tau}$ and ${\bf x}_t$. $-3 \leq \tau \leq 3$ is random in training, and $\tau=1$ in inference. 
Also, the reID feature is concatenated if available; if it is not, the neck feature is used instead (Fig. \ref{fig:detailview}).

\hspace{-6mm} {\bf 3: Clipped feature~$\tilde{{\bf q}}, \tilde{{\bf k}}, \tilde{{\bf v}}$.}
The input tokens ${\bf x}_t$ and ${\bf x}_{t-\tau}$ are then input to the cross-attention module. 
It first encodes the positional information to ${\bf x}_t$ and ${\bf x}_{t-\tau}$ ($pe(\cdot)$ in Eq. \ref{eq:pe}), as is also done in transformers~\cite{Ashish2017}. The output of $pe(\cdot)$ is denoted by ${\bf x}_{t, +}$ and ${\bf x}_{t-\tau, +}$. 
Next, ${\bf x}_{t, +}$, ${\bf x}_{t-\tau, +}$, and ${\bf x}_{t - \tau}$ are transformed into query~${\bf q}$, key~${\bf k}$, and value~${\bf v}$ (Eq. \ref{eq:def_embedding}), where $Q(\cdot)$, $K(\cdot)$ and $V(\cdot)$ are matrix multiplications.
In addition, we apply the \textit{feature clipping} ($clip(\cdot)$ in Eq. \ref{eq:feature_clipping}) to ${\bf q}$, ${\bf k}$, and ${\bf v}$, which reduces the number of elements (from 120 $\times$ 240 to 300).
The clipping position is the center of the bounding box. The clipped features are denoted by $\tilde{{\bf q}}$, $\tilde{{\bf k}}$, and $\tilde{{\bf v}}$.

\hspace{-6mm} {\bf 4: micro CNN~$\gamma(\cdot)$.}
We next extend $\tilde{{\bf q}}$ and $\tilde{{\bf k}}$ and obtain $\tilde{{\bf q}}'$ and $\tilde{{\bf k}}'$ by copying and concatenating their elements (Eq. \ref{eq:extend}) to form the token-wise pairs like the vanilla attention weight in transformers \cite{Ashish2017}.
The merged feature~$\tilde{{\bf q}}' \otimes \tilde{{\bf k}}'^{\top}$ is input to a micro CNN, denoted by $\gamma(\cdot)$ in Eq. \ref{eq:attention}, that is composed of two $1\times 1$ convolutions with the batch normalization and ReLU.
The micro CNN enriches the feature for the query-key process with the minimal cost.

\hspace{-6mm} {\bf 5: Cross-softmax function~$\phi(\cdot)$.}
The output of $\gamma(\cdot)$ is then input to the \textit{cross-softmax function} $\phi(\cdot)$ (Eq. \ref{eq:attention}).
$\phi(\cdot)$ estimates the \textit{affinity matrix} ${\bf A} \in \mathbb{R}^{N_{t} \times N_{t-\tau}}$, where $N_{t}$ and $N_{t-\tau}$ are the number of instances and tracklets, respectively. 
$\phi(\cdot)$ and ${\bf A}$ are defined as ${\bf A} = \phi({\bf r}) = {\bf r}_{cols} \otimes {\bf r}_{rows}$, where ${\bf r}_{cols} = \phi_{cols} ({\bf r})$ and ${\bf r}_{rows} = \phi_{rows} ({\bf r})$. $\phi_{cols}(\cdot)$ and $\phi_{rows}(\cdot)$ are the column- and row-softmax functions~\cite{Long2018} that normalize all the columns and rows of the output of $\gamma(\cdot)$ (${\bf r} = \gamma(\tilde{{\bf q}}' \otimes \tilde{{\bf k}}'^{\top})$), respectively.
The meaning of ${\bf A}$ is as follows: if ${\bf A}_{ij}$ is large, the $i_\mathrm{th}$ instance at $t$ and the $j_\mathrm{th}$ tracklet at $t-\tau$ are re-identified.
Importantly, ${\bf A}$ is a unimodal matrix \cite{Peixuan2022} and can be used for the one-to-one matching of instances and tracklets. As a result, TicrossNet no longer requires the Hungarian algorithm, which is widely used for the reID process but is time-consuming and non-differentiable.
Note that Time3D \cite{Peixuan2022} also computes the affinity matrix ${\bf A}$ for reID; however, they transform it with a matrix multiplication and also use the Hungarian algorithm. In contrast, we use ${\bf A}$ directly for reID and do not require the Hungarian algorithm.

To show the power of the proposed cross-softmax function, we compare the performance on the linear assignment problem.
Tab.~\ref{tab:ablation_assignment} shows that (1) a conventional method \cite{Long2018} cannot work well without the Hungarian algorithm, (2) despite this challenging setting (the large cost map), the cross-softmax function achieves 100$\%$ without the Hungarian algorithm, and (3) our cross-softmax function is faster than the conventional method with the Hungarian algorithm.

\begin{table}
  \centering
  \scalebox{1.0}{
  \begin{tabular}{c|cc}
    \hline 
    Method & Succ.~[$\%$] & Speed~[ms] \\
    \hline
    $\max({\bf r}_{cols}, {\bf r}_{rows})~\cite{Long2018}$ & 0.0 & 0.09 \\
    $\max({\bf r}_{cols}, {\bf r}_{rows})$ + H~\cite{Long2018} & 100.0 & 453.17 \\
    \hline
    ${\bf r}_{cols} \otimes {\bf r}_{rows}$ (ours) & 100.0 & 0.10 \\
    \hline
  \end{tabular}
  }
  \caption{The results of the linear assignment problem. ``H'' means the Hungarian algorithm. We test 10k trials using a $500 \times 500$ cost map, which is initialized as a normal distribution. SciPy \cite{scipy_hungarian} is used for H.}
  \label{tab:ablation_assignment}
\end{table}

\hspace{-6mm} {\bf 6: Output.}
Finally, Eq. \ref{eq:tix_attention} is applied.
$\beta (\cdot)$ inserts the clipped outputs (mentioned in {\bf 3: Clipped feature}) from $attention$ (Eq. \ref{eq:attention}) into the original position in the zero tensor~${\bf 0} \in O^{D \times (120 \times 240)}$.
The final outputs are ${\bf A}$ and ${{\bf x}'}_t$. ${\bf A}$ is used for re-identifying the instances at $t$ with the tracklets at $t-\tau$ (``ReID result'' in Fig. \ref{fig:detailview}), using the argmax operation to each column and thresholding. ${{\bf x}'}_t$ has exactly the same format as ${\bf x}_t$ (output from the detector) and now is refined. ${{\bf x}'}_t$ is used for finding bounding boxes (``Refined detection result'' in Figs. \ref{fig:overview} \& \ref{fig:detailview}).

\hspace{-6mm} {\bf 7: Loss function.}
The total loss is composed of five loss functions: $ L_{total} = \lambda_{t-\tau} L_{det}({\bf x}_{t-\tau}, {\bf y}_{det, t-\tau}) + \lambda_{t} L_{det}({\bf x}_{t}, {\bf y}_{det, t})$ $ + \lambda_{t} L_{det}({\bf x}'_{t}, {\bf y}_{det, t}) + e^{-\epsilon_{trk}} L_{trk}({\bf A}, {\bf y}_{a}) + \epsilon_{trk}$.
$L_{det}(\cdot)$ is the loss for the base detector, where ${\bf y}_{det, \cdot}$ is a ground truth label. $L_{det}(\cdot)$ is the same as that of CenteNet~\cite{Xingyi2019} in our case.
$L_{trk}(\cdot)$ is the loss for tracking and is the fast focal loss~\cite{Xingyi2020,Xingyi2019}, where ${\bf y}_a$ is a ground truth label and is an affinity-matrix-like binary matrix.
$\lambda_{\cdot}$ is a weight coefficient. 
We use $\lambda_{t-\tau}=0.5$ and $\lambda_{t}=0.25$.
If the reID feature is used (see {\bf 1: Detector} and {\bf 2: Input token ${\bf x}_t$}), we add a trainable coefficient~$\epsilon_{trk}$ \cite{Yifu2020}.

\subsection{Inference}\label{sec:loss}
{\bf 8: Track rebirth.}
To address the occlusion problem and re-identify the tracklets lost for a short time, we modify the track rebirth~\cite{Yifu2020,Xingyi2020,Nicolai2017} for TicrossNet.
The idea is simple: the affinity matrix ${\bf A}$ keeps a tracklet that is not re-identified with any instances at $t$, denoted by $j$. If $j$ does not reappear in 30 frames, $j$ is discarded. Note that our cross-attention mechanism successfully removes the attached modules for track rebirth, such as the boxIoU, Kalman filter, cosine similarity, and greedy matching \cite{Xingyi2020}, and therefore, we can finish all the key processes of tracking in an end-to-end MOT manner on GPU only, unlike the conventional models.

\hspace{-6mm} {\bf 9: Distance masking.}
To improve reID accuracy, we newly introduce \textit{distance masking} (different from \cite{Yifu2020}), which removes several irrelevant reID pairs; i.e., we ignore the reID pairs (in the sense of ${\bf A}$) that include an instance and a tracklet located at a distance.
Specifically, the distance masking inserts $-\infty$ to some elements in the output of micro CNN ${\bf r} = \gamma(\tilde{{\bf q}}' \otimes \tilde{{\bf k}}'^{\top})$. 
Such elements are selected if the distance between query~$i$ and key~$j$ (in the sense of the center of the bounding box) is larger than a threshold, where $i$ and $j$ are the index of an instance and a tracklet, respectively. The threshold is defined as $th_{i} = \min({\bf d}_{i}) + \alpha~\min(w_{i}, h_{i})$ ($\alpha=0.4$), where ${\bf d}_{i}$ is $ \{d(i, 1), d(i, 2), \ldots, d(i, j) \}$, $d(i, j)$ is the Euclidean distance between $i$ and $j$, and $w_{i}$ and $h_{i}$ are the width and height of the bounding box of $i$.
$-\infty$ after the softmax ensures that the reID pair does not contribute to ${\bf A}$.

\hspace{-6mm} {\bf 10: Memory sharing mechanism~(MSM).}
To reduce the cost of the detector and boost the inference speed, we modify MSM \cite{Jinlong2020}.
MSM for TicrossNet keeps the refined token ${\bf x}'_{t}$ in the memory and reuses it as ${\bf x}_{t'-\tau}$, where $t'=t+\tau$.

\section{Experiments}
\label{sec:exp}

{\bf Datasets.}
The test and training sets are as follows unless otherwise noted. The test set is two datasets for person tracking: test sets of MOT17~\cite{Anton2016} and MOT20~\cite{Patrick2020}.
The training set is composed of eight datasets~\cite{Yifu2020,Yifu2022}: Caltech~\cite{Piotr2012}, CityPerson~\cite{Shanshan2017}, ETHZ~\cite{Andreas2009}, CUHK-SYSU~\cite{Tong2016}, PRW~\cite{Liang2016}, MOT17 (training set), MOT20 (training set), and CrowdHuman~\cite{Shuai2018}. 
Training takes 4 days on 14 GPUs (Tesla V100).

\hspace{-6mm} {\bf Evaluation metrics.}
The standard CLEAR MOT metrics~\cite{Keni2008} are used: identity switch~(IDs), ID F1 score (IDF1), and MOT accuracy (MOTA).
The GPU used for TicrossNet is a single Tesla V100 except for Fig. \ref{fig:tradeoff_speed} \& Tab. \ref{tab:ablation_performance}.

\hspace{-6mm} {\bf Remark.} 
We acknowledge the importance of opening codes; however, our code includes a part of a commercial product, and thus we cannot publish it.

\begin{table}
  \centering
  \scalebox{1.0}{
  \begin{tabular}{c|ccccc}
    \hline 
    Tracker & MOTA~$\uparrow$ & IDF1~$\uparrow$ &  IDs~$\downarrow$ & FPS~$\uparrow$ \\
    \hline
    \multicolumn{5}{l}{MOT17} \\
    \hline
    DAN~(SST)~\cite{Long2018} & 52.4 & 49.5 & 8,431 & $<$~3.9 \\
    CenterTrack~\cite{Xingyi2020} & 67.8 & 59.9 & 2,898 & 22.0 \\
    TransCenter~\cite{Yihong2021}  & 73.2 & 62.2 & 4,614 & 1.0 \\
    FairMOT~\cite{Yifu2020}  & 73.7 & 72.3 & 3,303 & 25.9 \\
    TransTrack~\cite{Peize2020} & 75.2 & 63.5 & 3,603 & 10.0 \\
    MOTR~\cite{Zeng2022} & 73.4 & 68.6 & 2,439 & 7.5 \\
    ByteTrack~\cite{Yifu2022}  & {\bf 80.3} & {\bf 77.3} & {\bf 2,196} & 29.6 \\
    TicrossNet (ours) & 74.7 & 69.0 & 3,468 & {\bf 32.6} \\
    \hline
    \multicolumn{5}{l}{MOT20} \\
    \hline
    FairMOT~\cite{Yifu2020} & 61.8 & 67.3 & 5,243 & 13.2 \\
    TransCenter~\cite{Yihong2021} & 61.9 & 50.4 & 4,653 & 1.0 \\
    TransTrack~\cite{Peize2020} & 65.0 & 59.4 & 3,608 & 10.0 \\
    ByteTrack~\cite{Yifu2022}  & {\bf 77.8} & {\bf 75.2} & {\bf 1,223} & 17.5\\
    TicrossNet (ours) & 60.6 & 59.3 & 4,266 & {\bf 31.0} \\
    \hline
  \end{tabular}
  }
  \caption{Benchmark performance on MOT17 and MOT20 datasets. ``$<$'' means that detection speed is not included. The {\bf bold font} indicates the first-place performance on this leaderboard. Note that FPSs are the reported FPSs in each paper; thus, different GPUs are used for different models. A fair comparison is given in Fig. 3.}
  \label{tab:benchmark_mot}
\end{table}

\subsection{Benchmark Comparison \& Discussion}
Tab. \ref{tab:benchmark_mot} is a benchmark leaderboard on MOT17 and MOT20.
TicrossNet achieves 32.6 FPS on MOT17 and 31.0 FPS on MOT20 even though the latter includes as many as $>$100 instances per frame, while the other models including the state-of-the-art (SOTA) MOT model, ByteTrack~\cite{Yifu2022}, significantly slow down (except for the slow networks, i.e., TransCenter~\cite{Yihong2021} and TransTrack~\cite{Peize2020}).
Note that the video frame rates of MOT17 and MOT20 are 30 and 25 FPS, respectively; thus, we can safely say that TicrossNet runs \textit{in real-time}.
In terms of MOTA, IDF1, IDs, TicrossNet performs similarly to MOTR, which is the only end-to-end MOT model other than TicrossNet, but TicrossNet is significantly faster.

However, this high speed is at the expense of MOTA, IDF1, and IDs, compared with the other models. 
Let us discuss the reasons.
First, our empirical observation tells us that TicrossNet may not fully utilize motion information that may be captured by the Kalman filter in FairMOT. 
For example, two people walking side-by-side or crossing each other are difficult to track, and TicrossNet sometimes fails to track them, while FairMOT does not. Thus, appropriately involving motion information may improve the performance of TicrossNet.
Second, the performance gaps in MOTA, IDF1, and IDs between ByteTrack and TicrossNet partly come from the difference in the detectors: YOLOX \cite{Ge2021} and CenterNet, respectively.
In fact, our additional experiment shows that MOTA of TicrossNet with CenterNet and YOLOX on MOT17 half set \cite{Xingyi2020} is 62.6$\%$ and 68.4$\%$ (improved), respectively; however, that of ByteTrack is 75.8$\%$ \cite{Yifu2022} and still outperforms TicrossNet.
The gap may be closed if we use another base detector and/or adapt it to our pipeline.

\begin{figure}[t]
  \begin{center}
  \includegraphics[width=0.95\linewidth]{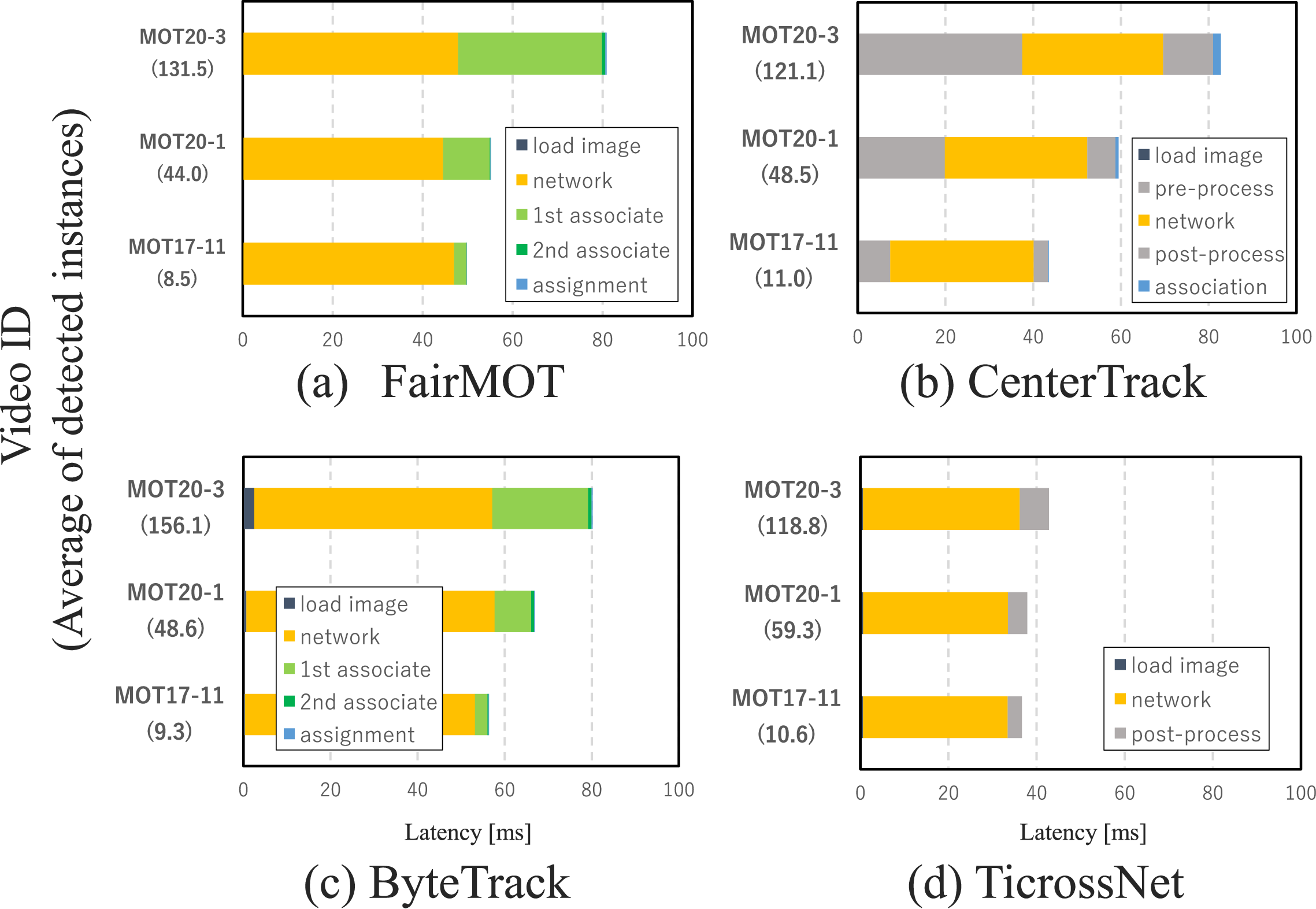}
  \caption{The number of instances vs. module latency on RTX 2080 Ti. The numbers in (*) mean the average numbers of detected instances per frame. MOT*-* are the video IDs. ``network'' (orange bars) means GPU processes. MOT17 half set \cite{Xingyi2020} is used.}  
  \label{fig:tradeoff_speed}
  \end{center}
\end{figure}

Nonetheless, in addition to the real-time speed, TicrossNet has an advantage that more than makes up for the low MOTA, IDF1, and IDs 
: \textit{the robustness to the number of instances} ($N_t$).
Fig. \ref{fig:tradeoff_speed} shows $N_t$ vs. module latency.
For fair comparison, the same GPU (RTX 2080 Ti) is used, while Tab. \ref{tab:benchmark_mot} is not. We pick out three fast models from Tab. \ref{tab:benchmark_mot}.
Fig. \ref{fig:tradeoff_speed} shows that the computational cost of TicrossNet does not increase significantly even when $N_t$ increases, unlike the other fast models including the SOTA MOT model, ByteTrack. 
This is because (1) TicrossNet can process all the key processes of MOT on GPU unlike the others, and (2) TicrossNet does not require the attached modules for tracking that tend to significantly increase computational cost when $N_t$ is large, as shown in Fig.\ref{fig:tradeoff_speed}.
Therefore, this result proves \textit{the robustness of TicrossNet to} $N_t$.
As a result, it does not have to change the size of the base detector, depending on $N_t$, as is often done by the other models.

\hspace{-6mm}  {\bf Ablation performance.}
Tab. \ref{tab:ablation_performance} shows an ablation study.
The cross-softmax dramatically improves MOTA, IDF1, IDs, and even FPS.
The micro CNN, reID feature, and distance masking also improve MOTA, IDF1, and IDs with minimal additional computational cost, as expected.

\begin{table}
  \centering
  \scalebox{0.85}{
  \begin{tabular}{l|cccc}
    \hline 
    Method & MOTA~$\uparrow$ & IDF1~$\uparrow$ & IDs~$\downarrow$ & FPS~$\uparrow$ \\
    \hline \hline
    Baseline            & -17.0 & 4.3 & 59.8 & 24.3 \\
    \hline
    + cross-softmax (par.~5)     & 46.4 & 47.1 & 1.3 & 26.3 \\
    + micro CNN (par.~4)         & 60.1 & 53.7 & 1.4 & 26.1 \\
    + reID feature (par.~1)      & 61.4 & 56.1 & 1.2 & 25.5 \\
    + distance masking (par.~9)  & 62.6 & 62.1 & 0.8 & 25.2 \\
    \hline
  \end{tabular}
  }
  \caption{Ablation study of TicrossNet on MOT17 half set~\cite{Xingyi2020} with RTX 2080 Ti. (par. *) means the paragraph number in Section \ref{sec:proposed}. ``Baseline'' is a modified TicrossNet (the micro CNN, reID feature, and distance masking are removed, and the cross-softmax is replaced with the counterpart of deep affinity network (DAN) \cite{Long2018}).
  } 
  \label{tab:ablation_performance}
\end{table}

\section{Conclusion}
\label{sec:conc}
TicrossNet is an end-to-end MOT model composed of a base detector and a single cross-attention module only and does not need any attached modules.
TicrossNet runs \textit{in real-time} even when the number of instances is $>$ 100.
Also, TicrossNet is robust to the change in the number of instances; thus, it does not have to change the size of the base detector, depending on the number of instances, as is often done by other models for real-time processing.

\vfill\pagebreak

\bibliographystyle{IEEEbib}
\bibliography{refs}

\end{document}